\documentclass{article} % For LaTeX2e
\usepackage{iclr2026_conference,times}

\usepackage{amsmath,amsfonts,bm}

\def\eqref#1{equation~\ref{#1}}
\def\1{\bm{1}}

\DeclareMathAlphabet{\mathsfit}{\encodingdefault}{\sfdefault}{m}{sl}
\SetMathAlphabet{\mathsfit}{bold}{\encodingdefault}{\sfdefault}{bx}{n}

\usepackage{hyperref}
\usepackage{url}
\usepackage{wrapfig}
\usepackage{algorithm}      % for the floating environment
\usepackage{algpseudocode}
\usepackage{pgfplots}
\pgfplotsset{compat=1.18}
\usepackage{multirow}
\usepackage{adjustbox}
\usepackage{arydshln}
\usepackage{colortbl}
\usepackage{multirow}
\usepackage[table]{xcolor}
\usepackage{booktabs}
\usepackage{tabularx}

\newcommand{\modelname}{\textsc{Compact}}
\newcommand{\commonact}{\textsc{Common-act$^2$}}
\newcommand{\actsquare}{\textsc{act$^2$}}
\newcommand{\interprune}{\textsc{FFN-pruning}}
\newcommand{\vocabprune}{\textsc{Vocab-pruning}}

\definecolor{lightred}{RGB}{255,127,127}
\definecolor{lightgreen}{RGB}{127,255,127}
\definecolor{lightblue}{RGB}{127,127,255}

\title{\modelname: Common-token Optimized Model Pruning Across Channels and Tokens}
\author{Eugene Kwek  and Wenpeng Yin \\
Department of Computer Science \& Engineering\\
Penn State University\\
\texttt{\{eyk5262,wenpeng\}@psu.edu} \\
}

\iclrfinalcopy % Uncomment for camera-ready version, but NOT for submission.
\begin{document}

\maketitle

\begin{abstract}
Making large language models (LLMs) more efficient in memory, latency, and serving cost is crucial for edge deployment, interactive applications, and sustainable inference at scale. Pruning is a promising technique, but existing pruning methods are limited: width pruning often breaks the standard transformer layout, requiring custom inference code, while depth pruning can cause abrupt accuracy drops. Also, while many pruning approaches are effective against LLMs, they struggle to maintain performance on small language models (SLMs). In this work, we propose \modelname, which jointly (i) prunes rare vocabulary to shrink embedding/LM head layers and (ii) prunes FFN intermediate channels using \emph{common-token–weighted} activations, aligning importance with the post-pruning token distribution. \modelname{} inherits strengths of both depth and width pruning, such as: deployment-friendliness (keeps a standard transformer architecture), scale-adaptivity (trade off vocab. vs. FFN pruning), competitive pruning times, and strong memory savings alongside throughput gains. Experiments across Qwen, LLaMA, and Gemma families (0.5B–70B) show state-of-the-art downstream performance, with substantial reductions in parameters, GPU memory, and latency\footnote{All code will be released, and the method will be packaged as a plug-in tool.}.
\end{abstract}

\section{Introduction}

Large language models (LLMs) have achieved remarkable performance across a wide range of natural language tasks, but their ever-growing parameter counts, reaching billions to hundreds of billions, make deployment expensive in terms of memory, inference time, and energy cost. To broaden access and enable real-world applications such as on-device inference, classroom use, or latency-sensitive systems, it is crucial to compress LLMs while retaining as much performance as possible.

Quantization \citep{DBLP:journals/corr/abs-2210-17323,DBLP:conf/mlsys/0002TTYCWXDG024} and pruning \citep{DBLP:conf/icml/FrantarA23,DBLP:conf/iclr/Sun0BK24} have been a major line of compression work. This work focuses on structured pruning, removing entire rows and columns of weight matrices. Structured pruning is mainly categorized into depth pruning and width pruning. Depth pruning removes entire transformer blocks~\citep{DBLP:journals/corr/abs-2402-02834,DBLP:conf/icml/SongOKKKK24,DBLP:conf/iclr/GromovTSGR25}, but the coarse-grained removal of layers leads to sharp performance drops. Width pruning trims hidden dimensions such as FFN channels or attention heads~\citep{DBLP:conf/nips/MaFW23,DBLP:conf/iclr/AshkboosCNHH24,DBLP:conf/aaai/AnZYTW24}, but they typically deviate from a standard transformer architecture and require custom inference code. In addition, these approaches are limited in three other ways:  
(i) They prune largely \emph{mechanistically}, without analyzing \emph{where parameters are concentrated} within LLMs (embeddings, FFNs, or attention). This blind pruning means that methods that work for large LLMs often fail for SLMs, as they have different parameter distributions.
(ii) They ignore the linguistic nature of NLP models: not all tokens are equally important, yet pruning typically treats all tokens as if they contribute equally. (iii) They often require custom implementation changes to accommodate every model family, making implementation maintenance tedious. These oversights lead to non-robust pruning performance across scales. 

To address these issues, we propose \modelname, a simple but powerful pruning framework with two modules:  
(i) \emph{Vocabulary pruning} removes rare tokens and shrinks embedding/LM head matrices, directly reducing parameters and memory usage, especially in SLMs.  
(ii) \emph{Common-token–weighted FFN pruning} further reduces redundancy by scoring channels using activations, but weighting only the common tokens that remain valid after vocabulary pruning.  
Together, these two complementary modules address the limitations of prior work: \textit{pruning is now guided by parameter distribution, respects the linguistic structure of language tasks, remains compatible with existing inference frameworks, and is architecture-agnostic across most model families}.

We systematically analyze parameter distributions across model families and scales. This reveals a clear pattern: embeddings (vocabulary and LM head layers) are important in SLMs, while FFNs dominate in larger models. This explains why prior pruning methods lack robustness across scales—they prune the same way regardless of where redundancy actually lies.  
A second insight comes from the statistics of natural language: token frequencies follow a Zipfian distribution \citep{DBLP:journals/corr/abs-2211-11041}, meaning that rare tokens occur extremely infrequently and contribute little to downstream performance. Removing such rare tokens from the vocabulary reduces embedding size without significantly affecting performance, because language tasks are overwhelmingly driven by common tokens. Together, these observations validate the effectiveness of the \modelname~method.

We evaluate \modelname{} on diverse LLM families (Qwen 2.5, LLaMA 3.1/3.2, and Gemma 3) and across scales from 0.5B to 70B parameters. We test on seven downstream benchmarks (MMLU \citep{DBLP:conf/iclr/HendrycksBBZMSS21}, HellaSwag \citep{DBLP:conf/acl/ZellersHBFC19}, WinoGrande \citep{DBLP:conf/aaai/SakaguchiBBC20}, ARC-C/E \citep{DBLP:journals/corr/abs-1803-05457}, PIQA \citep{DBLP:conf/aaai/BiskZLGC20}, GSM8K \citep{DBLP:journals/corr/abs-2110-14168}) and also measure pruning time, inference throughput, and GPU memory usage. Our experiments highlight three phenomena:
(i) \emph{Scale robustness:} \modelname{} maintains state-of-the-art performance at high pruning ratios even for SLMs.  
(ii) \emph{Smooth degradation:} Unlike depth pruning, which shows abrupt performance drops, \modelname{} degrades gracefully with higher pruning.
(iii) \emph{End-to-end efficiency:} \modelname~yields substantial GPU memory savings and improved throughput.

Our contributions are threefold: i) We provide a systematic analysis of parameter distribution across embeddings, FFNs, and attention, revealing scale-dependent redundancy that prior pruning methods overlook.  ii)  We propose \modelname{}, a novel pruning method which is linguistically grounded, scale-adaptive, and structure-agnostic.  iii) We demonstrate state-of-the-art pruning results across LLM families and scales, showing superior retention on downstream tasks together with clear gains in pruning time, inference efficiency, and GPU memory usage. 

\begin{table}[h]%[10]{r}{\textwidth}
\vspace{-5mm}
\caption{Advantages of \modelname.}
\begin{center}
\renewcommand{\arraystretch}{1.3}
\setlength{\tabcolsep}{4pt} % reduce horizontal padding
\small{
\begin{tabularx}{\textwidth}{l|ccc|ccc|c}
\toprule
& \multicolumn{3}{c|}{Depth pruning}& \multicolumn{3}{c|}{Width pruning}&  \modelname \\
& ShortGPT & LaCo & LLM-Streamline & SliceGPT & 2SSP & FLAP & (ours)\\
\midrule
%\cmidrule(lr){2-6}

%Type & Depth & Depth & Width & Width & \cellcolor{gray!20} Width \\
Maintains architecture & $\checkmark$ & $\checkmark$ & $\checkmark$ & & & & $\checkmark$ \\
Scale-adaptive & & & & & $\checkmark$ & & $\checkmark$ \\
Inference speedups & $\checkmark$ & $\checkmark$ & $\checkmark$ & $\checkmark$ & & $\checkmark$ & $\checkmark$ \\
Fast pruning & $\checkmark$ & $\checkmark$ & & $\checkmark$ & $\checkmark$ & $\checkmark$ & $\checkmark$ \\
Architecture-agnostic & & & & & & & $\checkmark$ \\
Linguistically grounded & & & & & & & $\checkmark$ \\
\bottomrule
\end{tabularx}
}\label{tab:advantages}
\end{center}
\end{table}
\vspace{-5mm}

\section{Related Work}

\textbf{Depth Pruning}
removes entire transformer blocks while preserving the standard architecture and compatibility with common inference frameworks \citep{DBLP:journals/corr/abs-2406-15786,DBLP:journals/corr/abs-2411-15558}. Representative methods include Shortened LLaMA (perplexity-minimizing), SLEB (iterative recalibration), and angular-similarity pruning \citep{DBLP:journals/corr/abs-2402-02834,DBLP:conf/icml/SongOKKKK24,DBLP:conf/iclr/GromovTSGR25}. LLM-Streamline trains a lightweight network to recover accuracy but requires hours–days and significant GPUs \citep{DBLP:conf/iclr/ChenHZWL025}. We therefore focus on training-free pruning that runs in minutes on a single GPU; \modelname{} can optionally be fine-tuned and outperforms training-based baselines. Because depth pruning is coarse-grained and can cause sharp drops, \modelname{} instead prunes rows/columns for a finer-grained alternative.

\textbf{Width Pruning}
removes hidden dimensions or channels in each layer \citep{DBLP:conf/iclr/XiaGZ024,DBLP:conf/nips/GaoLHTSJH24,DBLP:journals/corr/abs-2505-22689}. Methods include  LLM-Pruner/LoRAPrune (gradient-based) \citep{DBLP:conf/nips/MaFW23,DBLP:conf/acl/Zhang0SYOYZ24}, SliceGPT (orthogonal transforms + low-rank) \citep{DBLP:conf/iclr/AshkboosCNHH24}, FLAP (stability-based) \citep{DBLP:conf/aaai/AnZYTW24}, and Bonsai (perturbation modeling) \citep{DBLP:journals/corr/abs-2402-05406}. While effective, they often break the standard transformer layout, requiring custom inference code and limiting deployment. \modelname{} avoids these issues by preserving architecture, yielding a deployment-friendly width-pruning method that outperforms depth pruning.

% \paragraph{Width Pruning.}
% Width pruning \citep{DBLP:conf/iclr/XiaGZ024,DBLP:conf/nips/GaoLHTSJH24,DBLP:journals/corr/abs-2505-22689} often prunes each transformer layer in the LLM.  SliceGPT \citep{DBLP:conf/iclr/AshkboosCNHH24} prunes the hidden size by performing orthogonal matrix transformations and low-rank decomposition. LLM-Pruner \citep{DBLP:conf/nips/MaFW23} and LoRAPrune \citep{DBLP:conf/acl/Zhang0SYOYZ24} identifies and removes coupled structures using gradient information. FLAP \citep{DBLP:conf/aaai/AnZYTW24} prunes FFN channels and self-attention heads based on a channel stability metric. Bonsai \citep{DBLP:journals/corr/abs-2402-05406} builds a linear perturbation model of module importances, at the cost of significantly increased pruning time. Recent works \citep{DBLP:conf/icml/0006W0HWJLJPLBW24,DBLP:conf/coling/0002MWCS0XL025} have shown that non-uniform sparsity can improve pruning performance. However, these width pruning methods introduce structural irregularities that deviate from the transformer architecture. This makes it impractical to deploy width-pruned models, as they require specialized code for inference and are not compatible with standard inference libraries. \modelname~fixes these issues by ensuring that pruned models maintain a standard transformer architecture, enabling a deployment-friendly width-pruning method that outperforms depth pruning techniques.

\vspace{-3mm}
\paragraph{Vocabulary Size/Pruning.}

The vocabulary size of a model is the number of tokens that the model can recognize. Modern LLM vocabularies often reach into the hundreds of thousands and typically remain the same across model scales within a family \citep{DBLP:conf/nips/TaoLDMWLLW24}. Research into vocabulary size scaling \citep{DBLP:conf/nips/TaoLDMWLLW24} has shown that the optimal vocabulary size increases with increasing LLM size, contradicting the common practice of keeping vocabulary size constant over a wide range of model sizes. Prior work prunes vocabularies to tailor vocabulary to a target language/domain \citep{DBLPUshioZC23,DBLP02631,DBLP:conf/emnlp/UshioZC23,bogoychev2024upsdownslargelanguage,DBLPYangCC22,DBLP:journals/corr/abs-2412-15921}. Others prune the drafter’s LM head for speculative decoding speedups \citep{DBLP:conf/acl/ZhaoPHZAHZZLZZW25,DBLP:journals/corr/abs-2506-22694}, which does not compress the base model used at inference. In contrast, we (i) perform non-domain-specific vocabulary pruning for general-purpose LLMs, and (ii) couple it with common-token–weighted FFN pruning, so channel scores reflect the token distribution after vocab removal. This keeps a standard Transformer layout, is training-free, and proves robust from 0.5B to 70B.

\section{Proposed Method: \modelname}
Before designing an effective pruning strategy (Sections \ref{sec:vocab}-\ref{sec:joint} ), we first analyze where parameters are concentrated within modern decoder-only transformers (Section \ref{sec:distribution}).

\subsection{Analyzing Parameter Distribution in LLMs: \emph{Vocabulary} vs. \emph{FFN} vs. \emph{Attention}}\label{sec:distribution}

\begin{wrapfigure}[17]{r}{0.43\textwidth}
\vspace{-7mm}
\begin{center}
\begin{tikzpicture}
\begin{axis}[
    ybar stacked,
    bar width=10pt,
    width=0.42\textwidth,
    height=0.35\textwidth,
    ymin=0, ymax=1,
    ylabel={Proportion},
    symbolic x coords={0.5B,1.5B,3B,7B,14B,32B,72B},
    xtick=data,
    x tick label style={rotate=45,anchor=east},
    legend style={at={(0.5,-0.25)},anchor=north,legend columns=-1},
    enlarge x limits=0.15,
    axis line style={draw=none},
    ymajorgrids=true,
    grid style=dashed,
]
\addplot+[fill=lightblue] coordinates {(0.5B,0.276) (1.5B,0.151) (3B,0.101) (7B,0.143) (14B,0.105) (32B,0.048) (72B,0.034)};
\addplot+[fill=lightred] coordinates {(0.5B,0.635) (1.5B,0.749) (3B,0.789) (7B,0.749) (14B,0.690) (32B,0.830) (72B,0.800)};
\addplot+[fill=lightgreen] coordinates {(0.5B,0.089) (1.5B,0.100) (3B,0.110) (7B,0.108) (14B,0.204) (32B,0.123) (72B,0.166)};
\legend{Vocab,FFN,Attention}
\end{axis}
\end{tikzpicture}
\end{center}
\vspace{-2mm}
\caption{Parameter distribution across Qwen 2.5 models of different scales.}
\label{fig:distribution}
\end{wrapfigure}
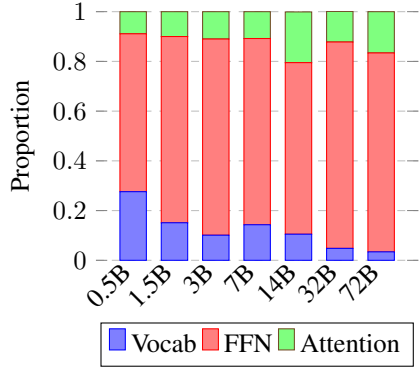
 Mainstream generative LLMs consist of three major groups of parameters:  
(i) \textit{vocab parameters}, located in the embedding and LM head layers;  
(ii) \textit{attention parameters}, from the self-attention blocks; and  
(iii) \textit{FFN parameters}, from the feed-forward blocks.

Formally, the embedding and LM head layers map between the vocabulary space of size $V$ and the hidden dimension $D$, giving
\begin{equation}
N_{\text{vocab}} = 2VD,
\end{equation}
(or $VD$ if tied embeddings are used). Each FFN block contains three projection matrices of size $D \times I$, where $I$ is the intermediate dimension, yielding
\begin{equation}
N_{\text{FFN}} = 3LDI,
\end{equation}
for $L$ layers. For attention, the number of parameters depends on whether grouped query attention  is used \citep{DBLP:conf/emnlp/AinslieLJZLS23}. With $H$ denoting the ratio of attention heads to KV heads, the count is
\begin{equation}
N_{\text{attention}} = 2LD^2\left(1+\frac{1}{H}\right).
\end{equation}
When $H=1$, this reduces to $N_{\text{attention}}=4LD^2$.

Asymptotically, $N_{\text{FFN}}$ and $N_{\text{attention}}$ scale as $O(LD^2)$—$I\approx O(D)$—while $N_{\text{vocab}}$ scales only as $O(D)$, as $V$ is kept constant when scaling. Thus, as model size grows, vocab parameters become proportionally smaller. Conversely, for smaller models, vocab parameters can constitute a significant fraction of the total. We observe this empirically by calculating the relative proportions of each parameter group on popular model families. Figure \ref{fig:distribution} shows our empirical analysis on the Qwen 2.5 model family \citep{DBLP:journals/corr/abs-2412-15115}, which validates our theoretical analysis. Proportions of other model families can be found in Appendix \ref{appendix:distribution}. 
This motivates our strategy: \textbf{vocabulary pruning is an efficient way to reduce parameters, especially in small-to-medium LLMs, while FFN pruning is critical for large models}.

\subsection{From Rare to Common: Rationale of Vocabulary Pruning}\label{sec:vocab}
\vspace{-2mm}
Byte-Pair Encoding (BPE) tokenizers follow Zipf's law \citep{DBLP:journals/corr/abs-2211-11041}, where most tokens appear extremely rarely. Since BPE builds its vocabulary by merging frequent token pairs, the rarest tokens naturally appear at the end of the vocabulary list.  

We define the set $S$ as the $V-V'$ rarest tokens in the vocabulary. These tokens can be directly removed by pruning the corresponding rows in the embedding/LM head matrices and deleting the corresponding merge rules from the tokenizer. The key insight is conceptual: \textbf{the deleted tokens will never be generated in the pruned model.} This means that subsequent optimization steps should focus on preserving performance under the \emph{common-token distribution} rather than the full distribution. \vocabprune~is highly efficient: it requires no calibration data or forward passes.

\subsection{Intermediate Pruning under the Common-Token Distribution}\label{sec:ffn}
\vspace{-2mm}
Pruning vocabulary parameters alone reduces the embedding size but does not address redundancies in the FFNs, which dominate parameter count in large models. To prune FFNs, we adopt an activation-based criterion. The standard \textit{act$^2$} method \citep{DBLP:conf/nips/MuralidharanSJC24,DBLP:journals/corr/abs-2501-17771} defines the importance of FFN intermediate channel $k$ as
\begin{equation}
\mathcal{I}_k = \sum_{i=1}^N \bigl(\text{SiLU}(X_iW_{\text{gate}})X_iW_{\text{up}}\bigr)_k^2,
\end{equation}
summing squared activations over a calibration dataset. Here, $X_i$ is FFN input and $W_{\text{gate}},W_{\text{up}}$ are model weights. However, this equally weights all tokens $x_i$, including $x_i\in S$. Since such tokens will never appear in the input after pruning, their activations should not guide channel importance. We therefore introduce \emph{common act$^2$}, a weighted variant:
\begin{equation}\label{eq:importance}
\mathcal{I}_k = \sum_{i=1}^Nw_i \bigl(\text{SiLU}(X_iW_{\text{gate}})X_iW_{\text{up}}\bigr)_k^2,
\quad
w_i = \begin{cases}
0 & x_i \in S, \\
1 & \text{otherwise}.
\end{cases}
\end{equation}
This ensures that FFN pruning is explicitly optimized for the tokens that remain valid after pruning.

\subsection{\modelname: Joint Pruning Pipeline}\label{sec:joint}
\vspace{-2mm}
Our proposed method, \modelname, integrates vocabulary pruning with common act$^2$-based FFN pruning. Importantly, embedding pruning and channel pruning are not performed sequentially in isolation: knowledge of $S$ (rarest tokens) is first identified, then used to guide intermediate pruning, and finally both vocab and FFN parameters are removed. The full pipeline is given in Algorithm~\ref{alg:compact}.

\begin{algorithm}[t]
\caption{\modelname}
\label{alg:compact}
\begin{algorithmic}[1]
\Require Model $M$, calibration dataset $\mathcal{D}$, target vocabulary size $V'$, target intermediate size $I'$
\State Identify $S \gets$ set of $V-V'$ rarest tokens in vocabulary.
\State Run forward passes of $M$ on $\mathcal{D}$, collect squared activations.
\State For each channel $k$, compute importance $\mathcal{I}_k$ using common act$^2$ (Eq. \ref{eq:importance}).
\For{each layer}
    \State Prune $I-I'$ least important channels (remove rows of $W_{\text{gate}}, W_{\text{up}}$ and columns of $W_\text{down}$).
\EndFor
\State Prune vocab parameters: remove final $V-V'$ rows of embedding and LM head matrices; delete tokenizer merges for tokens in $S$.
\State \Return pruned model $M'$.
\end{algorithmic}
\end{algorithm}

\textbf{Advantages of \modelname.}  
i) \textit{\modelname~is scale-adaptive.}
\modelname~uses two different knobs for pruning: (i) vocabulary pruning at the embedding/LM head layers and (ii) common-token–weighted pruning of FFN intermediate channels. These two knobs are orthogonal, allowing \modelname~to be tunable for SLMs (emphasize vocab pruning) LLMs (emphasize intermediate pruning), or any mix to meet a target budget. This tunability preserves capacity on frequent tokens while enabling strong compression across model scales.
ii) \textit{\modelname~is compatible with LLM frameworks.} One weakness of most width pruning methods is that they do not maintain a standard transformer architecture. This is indeed the case with SliceGPT, which prunes the hidden size in all layers except for the final one, as well as 2SSP, which prunes entire attention modules. As a result, these methods are not compatible with the transformers library, vLLM, or any other inference engines, limiting practicality. In this aspect, \modelname~is similar to depth pruning, since pruning the vocabulary and intermediate size does not affect the transformer architecture. As a result, \modelname~models are compatible with all inference engines, making it a practical approach to width pruning. The full advantages of \modelname~is summarized in Table \ref{tab:advantages}.

\section{Experiments}

\subsection{Experimental Setup}

\textbf{Baselines.} We compare with representative and state-of-the-art i) width pruning methods: \textit{SliceGPT}, \citep{DBLP:conf/iclr/AshkboosCNHH24}, \textit{2SSP} \citep{DBLP:journals/corr/abs-2501-17771}; ii) depth pruning methods: \textit{ShortGPT} \citep{DBLP:journals/corr/abs-2403-03853}, \textit{LaCo} \citep{DBLP:conf/emnlp/YangC024}. All methods use the default calibration set in their paper. For \modelname, we use 256 calibration samples from the C4 dataset \citep{DBLP:journals/jmlr/RaffelSRLNMZLL20}.

\textbf{LLMs to prune.}
We evaluate on a diverse set of LLMs spanning architectures and scales:
(i) \textbf{SLMs}: \textit{Qwen 2.5--0.5B}, \textit{LLaMA 3.2--1B}, and \textit{Gemma 3--1B}. 
This mix covers three distinct families, enabling a robustness assessment across architectures; moreover, small LLMs are particularly challenging to prune, as they are often trained beyond the Chinchilla‐optimal compute--data balance, leaving limited redundancy. \textbf{Nevertheless, pruning small LLMs is highly valuable for edge/on-device use and in privacy- or bandwidth-constrained settings (healthcare, classrooms, federated clients): it shrinks memory/storage, improves end-to-end latency, lowers energy use, and reduces serving cost.}
(ii) \textbf{LLMs}: \textit{LLaMA 3.1--8B} and \textit{LLaMA 3.1--70B}. 
Together with the 1B variant above, this suite evaluates pruning effectiveness across a wide scale.

\textbf{Evaluation tasks.} Following \textit{SliceGPT} \citep{DBLP:conf/iclr/AshkboosCNHH24}, we evaluate pruned models using \textbf{HellaSwag (HeSw)}, \textbf{WinoGrande (WiGr)}, \textbf{ARC-C}, \textbf{ARC-E}, and \textbf{PIQA}. Since \citet{DBLP:conf/iclr/JaiswalGDZWY24} shows that pruned LLMs degrade more on complex tasks, we add \textbf{MMLU} for general knowledge, and \textbf{GSM8K} for generation tasks. This gives a more complete view of model performance.

\textbf{Evaluation Criteria.} Details about the evaluation setup and pruning hyperparameters can be found in Appendix \ref{appendix:eval} and Appendix \ref{appendix:hyperparameters}, respectively. (A) We report the percentage of parameters that were removed \textbf{(Ratio (\%))}, the mean score of the 7 benchmarks \textbf{(Avg)}, and the relative mean score compared to the dense model \textbf{(Avg\%)}; (B) It is standard to evaluate pruned models on perplexity and downstream tasks. However, because we reduce vocabulary size, our perplexity naturally decreases, making comparisons to baselines unfair. Thus, we only report i) \textbf{performance on downstream tasks}, ii) \textbf{efficiency regarding pruning time, inference time, and memory usage}.

\subsection{Performance on downstream tasks}

\begin{table}[t]
\vspace{-4mm}
\caption{Pruning SLMs (Qwen 2.5-0.5B, LLaMA 3.2--1B, and Gemma 3--1B) at a $\sim$10\%, $\sim$20\%, and $\sim$35\% ratio. \textbf{Please note pruning baselines cannot be applied to Gemma 3}}
\vspace{-3mm}
\begin{center}
\begin{adjustbox}{width=\textwidth}
\begin{tabular}{c|cc|ccccccc|cc}
\hline
& \textbf{Method} & \textbf{Ratio (\%)} & \textbf{MMLU} & \textbf{HeSw} & \textbf{WiGr} & \textbf{ARC-C} & \textbf{ARC-E} & \textbf{PIQA} & \textbf{GSM8K} & \textbf{Avg} & \textbf{Avg\%} \\
\hline

\multirow{17}{*}{\rotatebox[origin=c]{90}{\textbf{Qwen 2.5--0.5B}}} &
Dense     & 0.00 & 47.3 & 52.2 & 56.4 & 32.3 & 58.2 & 69.9 & 34.9 & 50.2 & 100.0 \\
& Random    & -      & 25.0 & 25.0 & 50.0 & 25.0 & 25.0 & 50.0 & 0.0  & 28.6 & 57.0 \\
\cdashline{2-12}

& ShortGPT   & 9.11  & 27.8 & 44.0 & 53.0 & 25.9 & 45.8 & 66.8 & 0.2  & 37.6 & 75.1 \\
& LaCo       & 9.11  & \textbf{46.1} & 45.5 & \textbf{56.3} & 28.2 & 51.6 & 65.5 & 0.4  & 41.9 & 83.6 \\
& SliceGPT   & 10.71  & 23.2 & 43.2 & 53.3 & 26.4 & 50.3 & 63.8 & 0.0  & 37.2 & 74.1 \\
& 2SSP       & 10.12 & 25.5 & 46.6 & 54.7 & 27.8 & 52.2 & 68.7 & 1.9 & 39.6 & 79.0 \\
& \modelname & 11.13 & 45.2 & \textbf{51.9} & 55.3 & \textbf{32.4} & \textbf{59.5} & \textbf{70.1} & \textbf{28.7} & \textbf{49.0} & \textbf{97.7} \\
\cdashline{2-12}

& ShortGPT   & 18.02 & 25.0 & 37.7 & 52.0 & 27.1 & 41.7 & 62.1 & 0.0  & 35.1 & 70.0 \\
& LaCo       & 18.02 & 24.0 & 36.3 & 49.9 & 23.5 & 41.7 & 62.9 & 0.0  & 34.0 & 67.8 \\
& SliceGPT   & 19.64  & 23.1 & 32.1 & 52.0 & 20.2 & 33.4 & 53.7 & 0.0  & 30.6 & 61.1 \\
& 2SSP       & 19.64 & 24.3 & 40.9 & 53.8 & 25.3 & 43.9 & 64.3 & 0.5 & 36.1 & 72.0 \\
& \modelname & 20.24 & \textbf{44.1} & \textbf{48.1} & \textbf{55.4} & \textbf{30.6} & \textbf{53.3} & \textbf{66.6} & \textbf{26.3} & \textbf{46.3} & \textbf{92.4} \\
\cdashline{2-12}

& ShortGPT   & 36.23 & 24.3 & 27.8 & 50.1 & \textbf{25.7} & 26.2 & 51.8 & 0.0 & 29.4 & 58.7 \\
& LaCo       & 36.23 & 23.9 & 28.2 & 47.9 & 23.9 & 30.6 & 56.2 & 0.0 & 30.1 & 60.0 \\
& SliceGPT   & 36.61 & 23.1 & 29.0 & 51.5 & 22.5 & 30.7 & 53.4 & 0.0 & 30.0 & 59.9 \\
& 2SSP       & 36.23 & 22.9 & 31.3 & 49.6 & 22.9 & 33.8 & 59.0 & 0.0 & 31.4 & 62.5 \\
& \modelname & 37.04 & \textbf{25.5} & \textbf{40.0} & \textbf{53.8} & 25.2 & \textbf{40.0} & \textbf{62.2} & \textbf{0.5} & \textbf{35.3} & \textbf{70.4} \\
\hline

%llama3.2 start

\multirow{17}{*}{\rotatebox[origin=c]{90}{\textbf{LLaMA 3.2--1B}}}& Dense     & 0.00 & 36.6 & 63.8 & 60.7 & 36.2 & 60.7 & 74.5 & 5.4 & 48.3 & 100.0 \\
&Random    & -      & 25.0 & 25.0 & 50.0 & 25.0 & 25.0 & 50.0 & 0.0 & 28.6 & 59.2 \\
\cdashline{2-12}

&ShortGPT    & 10.03 & 23.4 & 48.0 & \textbf{60.2} & 30.4 & 49.8 & 68.0 & 0.0  & 40.0 & 82.8 \\
&LaCo        & 10.03 & 24.4 & 48.4 & 52.0 & 29.5 & 47.9 & 69.3 & 0.5  & 38.9 & 80.5 \\
&SliceGPT    & 10.16 & 23.1 & 49.4 & 54.0 & 28.6 & 43.4 & 64.0 & 0.0  & 37.5 & 77.7 \\
&2SSP        & 9.87  & 30.5 & 55.5 & 57.9 & 32.9 & 56.7 & \textbf{72.9} & 3.0 & 44.2 & 91.6 \\
&\modelname  & 10.03 & \textbf{36.7} & \textbf{61.1} & 59.7 & \textbf{35.1} & \textbf{57.1} & 71.9 &  \textbf{6.1}  & \textbf{46.8} & \textbf{97.0} \\
\cdashline{2-12}

&ShortGPT   & 19.66 & 22.8 & 40.0 & 55.3 & 29.9 & 35.2 & 58.9 & 0.0  & 34.6 & 71.7 \\
&LaCo       & 19.66 & 23.0 & 35.7 & 52.4 & 25.9 & 37.0 & 62.4 & 0.4  & 33.9 & 70.1 \\
&SliceGPT   & 20.31 & 23.0 & 39.9 & 52.3 & 26.2 & 38.9 & 58.6 & 0.0  & 34.1 & 70.7 \\
&2SSP       & 19.66 & 26.8 & 46.9 & 53.9 & 27.1 & 50.4 & 68.1 & 2.2  & 39.3 & 81.5 \\
&\modelname & 19.98 & \textbf{30.6} & \textbf{54.4} & \textbf{58.6} & \textbf{32.0} & \textbf{51.3} & \textbf{69.9} &   \textbf{3.1}  & \textbf{42.8} & \textbf{88.8} \\
\cdashline{2-12}

&ShortGPT   & 34.47 & 24.3 & 32.5 & 50.0 & \textbf{28.4} & 28.9 & 55.4 & 0.0 & 31.4 & 65.0 \\
&LaCo       & 34.47 & 23.2 & 37.3 & 50.1 & 23.9 & 29.3 & 55.3 & 0.0 & 29.9 & 61.9 \\
&SliceGPT   & 35.16 & 23.0 & 30.4 & 51.2 & 22.0 & 32.9 & 53.4 & 0.0 & 30.4 & 63.1 \\
&2SSP       & 34.63 & 22.9 & 35.1 & 52.6 & 24.5 & 38.9 & \textbf{60.9} & 0.0 & 33.6 & 69.6 \\
&\modelname & 35.03 & \textbf{27.9} & \textbf{42.8} & \textbf{55.6} & 27.7 & \textbf{41.7} & 60.6 &   \textbf{1.8} & \textbf{36.9} & \textbf{76.4} \\
\hline

% Gemma starts
\multirow{5}{*}{\rotatebox[origin=c]{90}{\textbf{Gemma3--1B}}}& Dense & 0.00 & 24.9 & 62.1 & 59.0 & 38.2 & 71.9 & 74.8 & 2.4 & 47.6 & 100.0 \\
&Random & -      & 25.0 & 25.0 & 50.0 & 25.0 & 25.0 & 50.0 & 0.0 & 28.6 & 60.0 \\
\cdashline{2-12}

&\multirow{3}{*}{\modelname} & 10.01 & 24.9 & 60.3 & 59.0 & 39.1 & 69.0 & 74.1 & 1.7 & 46.9 & 98.4 \\
&      & 20.02 & 25.0 & 55.4 & 59.0 & 37.9 & 63.3 & 70.0 & 1.7 & 44.6 & 93.6 \\
&      & 34.99 & 24.2 & 45.1 & 55.9 & 26.5 & 46.4 & 65.3 & 0.5 & 37.7 & 79.2 \\
\hline
\end{tabular}\label{tab:sllm}
\end{adjustbox}
\end{center}
\vspace{-5mm}
\end{table}

\subsubsection{Results on Smaller LLMs}

\paragraph{\modelname~outperforms baselines by large margins.} Our results are summarized in Table \ref{tab:sllm}. Although prior works report strong results on LLMs, they perform poorly on SLMs. On Qwen 2.5--0.5B, GSM8K accuracy collapses for all baselines at only 10\% pruning. Likewise, at 10\% pruning, MMLU drops to near-random for all models and baselines, with the sole exception of LaCo on Qwen 2.5--0.5B. Because MMLU and GSM8K are the most demanding tasks in our suite, these trends indicate that existing methods fail to preserve performance on challenging benchmarks for SLMs.
In contrast, \modelname~delays this collapse: it remains marginally above random on MMLU and GSM8K even at 35\% pruning across all models. Across models and pruning ratios, \modelname~attains the highest mean score and leads on nearly all individual benchmarks, despite using a slightly higher pruning ratio than the baselines. Notably, at 35\% pruning, \modelname~maintains similar performance to the baselines at 20\%, demonstrating superior robustness under high compression.

\paragraph{\modelname~supports a wide variety of model architectures out-of-the-box.} The official implementations of existing approaches support just a few model architectures. For instance, SliceGPT supports LLaMA/OPT/Phi, LaCo supports LLaMA 2/Baichuan, and ShortGPT only supports LLaMA. Adding support for modern model families like LLaMA 3 and Qwen 2.5 required adding architecture-specific changes, since these architectures can differ significantly in how they handle self-attention, layer normalization, etc. The strength of \modelname~is that it only prunes the vocabulary embeddings and FFN blocks, which have been standardized and remain unchanged across the vast majority of model architectures. As a consequence, \modelname~is architecture-agnostic and runs out-of-the-box across many model families. This is most evident with Gemma~3, which uses QK-norm and alternating local and global attention layers—optimizations that prevented us from adapting our baselines. Accordingly, baseline results for Gemma~3 are omitted in Table~\ref{tab:sllm}. In contrast, \modelname~operates on Gemma~3 without any architecture-specific changes.

\begin{table}[t]
\vspace{-3mm}
\caption{Pruning larger LLMs (LLaMA 3.1--8B \& LLaMA 3.1--70B) at a $\sim$10\%, $\sim$20\%, and $\sim$35\% ratio. LaCo failed to prune LLaMA 3.1-70B due to OOM errors, so it is omitted from our results.}
\begin{center}
\begin{adjustbox}{width=\textwidth}
\begin{tabular}{c|cc|ccccccc|cc}
\hline
& \textbf{Method} & \textbf{Ratio (\%)} & \textbf{MMLU} & \textbf{HeSw} & \textbf{WiGr} & \textbf{ARC-C} & \textbf{ARC-E} & \textbf{PIQA} & \textbf{GSM8K} & \textbf{Avg} & \textbf{Avg\%} \\
\hline

\multirow{17}{*}{\rotatebox[origin=c]{90}{\textbf{LLaMA 3.1--8B}}}& Dense & 0.00  & 63.4 & 78.9 & 73.6 & 53.4 & 80.9 & 81.1 & 51.6 & 69.0 & 100.0 \\
&Random & -       & 25.0 & 25.0 & 50.0 & 25.0 & 25.0 & 50.0 & 0.0  & 28.6 & 41.4 \\
\cdashline{2-12}

&ShortGPT   & 10.86 & 58.0 & 73.8 & 70.2 & 47.4 & 71.2 & 77.5 & 29.3 & 61.1 & 88.5 \\
&LaCo       & 10.86 & 58.8 & 73.3 & 72.3 & 48.9 & 73.7 & 76.3 & \textbf{32.0} & 62.2 & 90.1 \\
&SliceGPT   & 10.16 & 43.9 & 65.6 & 67.2 & 39.4 & 66.2 & 71.0 & 19.4 & 53.2 & 77.2 \\
&2SSP       & 10.86 & 54.1 & 74.6 & 71.6 & 46.8 & 71.6 & \textbf{79.7} & 14.1 & 58.9 & 85.4\\
&\modelname & 10.00 & \textbf{59.6} & \textbf{75.2} & \textbf{73.7} & \textbf{50.3} & \textbf{74.9} & 78.4 & 27.8 & \textbf{62.9} & \textbf{91.1} \\
\cdashline{2-12}

&ShortGPT   & 19.02 & \textbf{58.6} & 64.9 & 68.4 & 42.2 & 58.3 & 71.6 & 0.6  & 52.1 & 75.5 \\
&LaCo       & 19.02 & 24.1 & 54.0 & 55.3 & 29.2 & 51.1 & 72.4 & 0.4  & 40.9 & 59.3 \\
&SliceGPT   & 20.12 & 24.5 & 51.4 & 61.8 & 30.3 & 49.0 & 61.9 & 0.0  & 39.8 & 57.8 \\
&2SSP       & 19.99 & 37.4 & 67.2 & 68.4 & 38.1 & 61.8 & \textbf{76.8} & 4.3  & 50.6 & 73.3 \\
&\modelname & 20.00 & 50.7 & \textbf{69.9} & \textbf{70.1} & \textbf{42.8} & \textbf{66.0} & 75.9 & \textbf{10.8} & \textbf{55.2} & \textbf{80.0} \\

\cdashline{2-12}

&ShortGPT   & 35.31 & 23.2 & 34.3 & 59.1 & 29.7 & 36.9 & 57.2 & 0.0  & 34.4 & 49.8 \\
&LaCo       & 35.31 & 23.1 & 34.8 & 53.1 & 27.3 & 31.5 & 58.8 & 0.0  & 32.7 & 47.3 \\
&SliceGPT   & 35.16 & 23.0 & 35.0 & 54.3 & 23.5 & 37.2 & 55.0 & 0.0  & 32.6 & 47.2 \\
&2SSP       & 34.77 & 25.3 & 49.9 & 59.3 & 27.1 & 44.3 & 68.7 & \textbf{2.3}  & 39.5 & 57.3 \\
&\modelname & 34.99 & \textbf{35.9} & \textbf{56.0} & \textbf{63.3} & \textbf{30.8} & \textbf{48.4} & \textbf{70.6} & 1.7  & \textbf{43.8} & \textbf{63.5} \\
\hline

% 70B starts

\multirow{14}{*}{\rotatebox[origin=c]{90}{\textbf{LLaMA 3.1--70B}}}&Dense & 0.00  & 75.2 & 85.0 & 79.5 & 64.7 & 86.7 & 84.4 & 80.6 & 79.4 & 100.0 \\
&Random & -       & 25.0 & 25.0 & 50.0 & 25.0 & 25.0 & 50.0 & 0.0  & 28.6 & 36.0 \\
\cdashline{2-12}

&ShortGPT   & 9.77  & \textbf{75.0} & 82.9 & 78.5 & 60.8 & 84.8 & 83.4 & 74.5 & 77.1 & 97.1 \\
&SliceGPT   & 10.06 & 70.6 & 75.4 & 76.6 & 58.3 & 82.0 & 79.7 & 62.4 & 72.1 & 90.8 \\
&2SSP       & 9.92  & 73.4 & \textbf{84.7} & 78.6 & 63.9 & 85.4 & \textbf{84.2} & 75.1 & 77.9 & 98.1 \\
&\modelname & 10.06 & 73.5 & 84.5 & \textbf{79.5} & \textbf{64.0} & \textbf{86.0} & 84.1 & \textbf{76.0} & \textbf{78.2} & \textbf{98.5} \\
\cdashline{2-12}

&ShortGPT   & 20.68 & \textbf{74.9} & 79.3 & \textbf{78.0} & 56.0 & 80.1 & 80.1 & 53.0 & 71.6 & 90.2 \\
&SliceGPT   & 20.02 & 63.1 & 64.8 & 74.0 & 52.4 & 76.7 & 73.7 & 0.0  & 57.8 & 72.8 \\
&2SSP       & 20.11 & 68.8 & \textbf{83.7} & 76.2 & 59.5 & 82.1 & \textbf{83.7} & 62.1 & 73.7 & 92.8 \\
&\modelname & 20.11 & 70.6 & 83.3 & 76.2 & \textbf{59.7} & \textbf{82.6} & \textbf{83.7} & \textbf{62.6} & \textbf{74.1} & \textbf{93.3} \\
\cdashline{2-12}

&ShortGPT   & 36.40 & \textbf{71.2} & 66.5 & \textbf{75.7} & 46.8 & 69.6 & 72.3 & 0.0  & 57.4 & 72.3 \\
&SliceGPT   & 35.06 & 29.3 & 37.1 & 66.6 & 34.5 & 59.3 & 62.6 & 0.0  & 41.3 & 52.0 \\
&2SSP       & 35.13 & 58.5 & 77.7 & 72.1 & 50.7 & 74.2 & \textbf{81.6} & \textbf{25.0} & 62.8 & 79.1 \\
&\modelname & 34.99 & 59.6 & \textbf{78.1} & 72.8 & \textbf{53.2} & \textbf{76.1} & 81.3 & \textbf{25.0} & \textbf{63.7} & \textbf{80.2} \\
\hline
\end{tabular}\label{tab:largellms}
\end{adjustbox}
\end{center}
\vspace{-4mm}
\end{table}

\subsubsection{Results on larger LLMs}

\paragraph{\modelname~is robust across scales.} Our results are in Table \ref{tab:largellms}. We see that \modelname~achieves state-of-the-art performance for larger models as well, with over 80\% performance at a 35\% ratio, indicating that our method is highly robust to a wide range of sizes. We attribute this robustness to our dual approach to width pruning. As model size increases, the proportion of vocabulary parameters decreases, which decreases the effectiveness of pruning vocabulary size. However, intermediate pruning becomes more effective as model size increases, similarly to most pruning methods \citep{xu2024rethinking}. These two techniques complement each other's strengths, leading to a robust hybrid pruning method. Although the performance gap between \modelname~and 2SSP narrows at the 70B size, we note that 2SSP's hybrid approach of pruning FFN channels and entire attention blocks deviates from the standard transformer architecture, while \modelname's approach does not. This makes \modelname~more practical to deploy, while also achieving slightly higher performance.

\paragraph{\modelname~shows smooth degradation.} On LLaMA 3.1--8B, we observe that ShortGPT surpasses \modelname~at a 20\% pruning ratio, but not at 10\% or 35\%, with its 20\% score even exceeding its 10\% score. This aligns with the \emph{step-like} behavior in \citet{DBLP:conf/iclr/GromovTSGR25} for depth-pruning methods, where performance remains intact up to a critical threshold and then collapses abruptly. ShortGPT also exhibits this pattern, explaining the spike at 20\%. In contrast, \modelname~(a width-pruning method) shows smooth MMLU degradation as pruning increases—hence the dip relative to ShortGPT at 20\%, followed by recovery and a lead at 35\%. Even at 35\% pruning, \modelname~remains above random on both MMLU and GSM8K. A similar effect is seen on LLaMA 3.1--70B.

\begin{wraptable}[13]{r}{0.35\textwidth}
\vspace{-7mm}
\caption{Proportion of words retokenized after 35\% pruning of Qwen 2.5–0.5B, by dataset.}
\begin{center}
\begin{tabular}{c|c}
\hline
\textbf{Dataset} & \textbf{Rare\%} \\
\hline
\textbf{MMLU}        & 4.43\% \\
\textbf{HellaSwag}   & 3.48\% \\
\textbf{WinoGrande}  & 5.01\% \\
\textbf{ARC-C}       & 3.82\% \\
\textbf{ARC-E}       & 3.95\% \\
\textbf{PIQA}        & 5.68\% \\
\textbf{GSM8K}       & 3.60\% \\
\textbf{C4}          & 4.56\% \\
\hline

\end{tabular}\label{tab:rarevocab}
\end{center}
\end{wraptable}

\paragraph{Analysis: Why our pruning hurts performance minimally?}

Changing the vocabulary affects how text is tokenized: If a token is removed during pruning, the text associated with the token is now tokenized as multiple shorter, more common tokens. We analyze how often rare tokens occur. We use the questions in each of our benchmarks, as well as 10k random samples from the C4 dataset. Our results are in Table \ref{tab:rarevocab}. Our pruned model reduces vocabulary size from 150k to 50k, a 67\% reduction. Despite this, only $4\%$ of words are tokenized differently from the original, regardless of text source. These results provide insight to the effectiveness of \vocabprune: significant proportions of vocabulary only affects a small fraction of text, so removing these vocabulary has little impact on performance.

\subsection{Efficiency in Pruning Time, Inference Latency, and Memory}

\vspace{-3mm}
\paragraph{Pruning time.}
The main strength of training-free methods is that they can prune large models efficiently. To test this, we report pruning times at 35\% pruning on LLaMA 3.1-8B and 70B to best discriminate between methods.

\begin{wraptable}[15]{r}{0.35\textwidth}
\vspace{-4mm}
\caption{3-run average pruning time (mm:ss) comparison at a 35\% pruning ratio for LLaMA 3.1. We exclude I/O time for fairness.}
\vspace{-3mm}
\begin{center}
\begin{tabular}{c|c|c}
\hline
 & \textbf{Method} & \textbf{Pru. Time} \\
\hline

\multirow{5}{*}{8B}
& ShortGPT & 0:18 \\
& LaCo     & 0:05 \\
& SliceGPT & 10:48 \\
& 2SSP     & 1:26 \\
& \modelname & 0:32 \\
\hline

\multirow{5}{*}{70B}
& ShortGPT & 2:10 \\
& SliceGPT & 84:38 \\
& 2SSP     & 13:48 \\
& \modelname & 2:17 \\
\hline

\end{tabular}\label{tab:pruningtime}
\end{center}
\end{wraptable}
Our results are in Table \ref{tab:pruningtime}. With LLaMA 3.1-8B, \modelname~is 3 times faster than 2SSP, our strongest baseline, and comparable our depth pruning methods, with pruning times under a minute. At the 70B size, \modelname~now becomes 6 times faster than 2SSP. The low pruning times show that \modelname~has competitive efficiency to our baselines.

\vspace{-3mm}
\paragraph{Inference speed and memory usage.}

\begin{table}%[htbp]
\vspace{-3mm}
\caption{Throughput and memory usage for LLaMA 3.1-8B.}
\begin{center}
\begin{tabular}{c|cc|cc}
\hline
&\textbf{Method} & \textbf{Ratio} & \textbf{Memory Usage (MB)} & \textbf{Throughput (q/s)} \\
\hline

\multirow{4}{*}{Classification}& Dense          & 0.00\%  & 50030 & 147.01 \\
\cdashline{2-5}
& ShortGPT/LaCo  & 35.31\% & 44624 (0.89x) & \textbf{221.61 (1.51x)} \\
& 2SSP           & 34.77\% & 44985 (0.90x) & 104.03 (0.71x) \\
& SliceGPT       & 35.16\% & 42440 (0.85x) & 173.94 (1.18x) \\
& \modelname           & 34.99\% & \textbf{32066 (0.64x)} & 201.19 (1.37x) \\
\hline

%generation
\multirow{3}{*}{Generation}& Dense          & 0.00\%  & 21787 & 81.18 \\
\cdashline{2-5}
& ShortGPT/LaCo  & 35.31\% & 16336 (0.75x)  & \textbf{128.03 (1.57x)} \\
& \modelname           & 34.99\% & \textbf{14248 (0.65x)} & 112.40 (1.38x) \\
\hline

\end{tabular}\label{tab:inference}
\end{center}
\vspace{-7mm}
\end{table}

We evaluate two inference paradigms: \emph{Text classification} and \emph{Text generation}. The inference setup is in Appendix \ref{appendix:inference}.

Our results are in Table \ref{tab:inference}. Note that ShortGPT and LaCo prune the same number of layers, so they have the same inference performance. In the text classification task, our method achieves the highest memory reduction, and the second-highest throughput increase. The low memory usage is from the reduced vocabulary size. During the forward pass, logits are stored in GPU memory, which becomes very large with high batch sizes. By pruning the vocabulary size, \modelname~shrinks logit size, causing large memory reductions. \textbf{This is especially important in edge computing applications, where memory is very limited and can spell the difference in whether a model can be used or not.} \modelname~has higher throughput than our width pruning baselines SliceGPT/2SSP, although it falls behind depth pruning methods. This aligns with \citet{DBLP:journals/corr/abs-2501-18107}, which showed that scaling down layer count proportionally increases throughput, but scaling layer size does not. While \modelname~achieves faster inference than other width pruning methods, more work is needed to match the throughput of depth pruning methods. In the generation task, the trends in memory usage and throughput are similar to that of the classification task.

\vspace{-2mm}
\subsection{Recovery Fine-tuning}
\vspace{-2mm}
Although \modelname{} is training-free, we optionally apply recovery fine-tuning. We fine-tune Qwen 2.5–0.5B pruned at 35\% using two approaches: (i) Continued Pretraining (CPT): train on 900M tokens from FineWeb-Edu \citep{DBLP:conf/nips/PenedoKALMRW024}; (ii) Self-Data Distillation (SDD): generate 900M tokens from the unpruned model (temperature = 1.0) to match its training distribution, then fine-tune on this synthetic data \citep{DBLP:journals/corr/abs-2410-09982}.
% Although \modelname~is a training-free method, we can optionally perform recovery fine-tuning to regain performance. To test this, we perform full finetuning on Qwen 2.5-0.5B after pruning at a 35\% pruning ratio. We conduct two experiments with this setup: (i) \textbf{Continued Pretraining (CPT):} We use the first 900 million tokens of the FineWeb-Edu dataset \citep{DBLP:conf/nips/PenedoKALMRW024}. (ii) \textbf{Self-Data Distillation (SDD):} Modern model families like Qwen use high-quality, private pretraining datasets that have a different distribution than open-source versions. To address this, we adopt a modified version of \textit{self-data distillation} \citep{DBLP:journals/corr/abs-2410-09982}, which creates a dataset with the same distribution as the training dataset. More specifically, we use the unpruned 0.5B model to generate 900 million tokens of text completions, seeded by the first token of documents in FineWeb-Edu. We use temperature=1.0 and no other sampling parameters to ensure that the dataset reflects the unmodified probability distribution of the unpruned model. We then fine-tune our pruned model on this synthetic dataset.

Despite some incoherence in the SDD synthetic dataset, our results (Table \ref{fig:finetuning}) show that SDD consistently improves all benchmarks and yields a higher average score than CPT, which fails to boost MMLU or GSM8K. This confirms SDD’s effectiveness even below 1B parameters \citep{DBLP:journals/corr/abs-2410-09982}. With fine-tuning, \modelname{} surpasses training-based LLM-Streamline \citep{DBLP:conf/iclr/ChenHZWL025} and even outperforms Gemma 3-270M, pretrained on 6T tokens—highlighting pruning’s efficiency relative to pretraining.

% Our results are in Table \ref{fig:finetuning}. Manually examining the SDD dataset, we find that many samples are incoherent, with repetitive text and hallucinations. Despite this, we find that SDD significantly outperforms CPT. Although CPT achieves higher scores on most benchmarks, it sees no improvement on MMLU or GSM8K, the two hardest benchmarks. SDD consistently improves on all benchmarks, enabling a higher average score. These results show that SDD outperforms CPT even at the sub-1B parameter range, extending the results in \citet{DBLP:journals/corr/abs-2410-09982}. With fine-tuning, our method significantly outperforms the training-based method LLM-Streamline \citep{DBLP:conf/iclr/ChenHZWL025}. Incredibly, our method fine-tuned with 900M tokens also outperforms the recent Gemma 3-270M, pretrained on 6 \emph{trillion} tokens. These results highlight the efficiency of pruning compared to pretraining.

\begin{table}%[htbp]
\caption{Comparison of post-pruning recovery fine-tuning methods.}
\vspace{-3mm}
\begin{center}
\begin{adjustbox}{width=\textwidth}
\begin{tabular}{c|c|ccccccc|cc}
\hline
\textbf{Method} & \textbf{Params (M)} & \textbf{MMLU} & \textbf{HeSw} & \textbf{WiGr} & \textbf{ARC-C} & \textbf{ARC-E} & \textbf{PIQA} & \textbf{GSM8K} & \textbf{Avg} & \textbf{Avg\%} \\
\hline

Dense          & 494  & 47.3 & 52.2 & 56.4 & 32.3 & 58.2 & 69.9 & 34.9 & 50.2 & 100.0 \\
Random         & -    & 25.0 & 25.0 & 50.0 & 25.0 & 25.0 & 50.0 & 0.0  & 28.6 & 57.0 \\
LLM-Streamline & 315  & 23.0 & 31.8 & 53.4 & 23.5 & 37.9 & 59.7 & 0.2 & 32.8 & 65.4 \\

Gemma 3-270M & 268 & 26.2 & 41.5 & 53.8 & 28.4 & \textbf{57.2} & \textbf{68.4} & 1.2 & 39.5 & - \\
\hline
Our \modelname           & 311 & 25.5 & 40.0 & 53.8 & 25.2 & 40.0 & 62.2 & 0.5  & 35.3 & 70.4 \\
\textit{+ CPT} & 311  & 25.8 & \textbf{45.1} & \textbf{55.8} & 27.9 & 52.7 & 67.7 & 0.0  & 39.3 & 78.3 \\
\textit{+ SDD} & 311  & \textbf{39.7} & 43.0 & 54.9 & \textbf{28.6} & 50.6 & 65.6 & \textbf{11.3} & \textbf{41.9} & \textbf{83.6} \\
\hline

\hline

\end{tabular}
\end{adjustbox}
\end{center}
\label{fig:finetuning}
\vspace{-4mm}
\end{table}

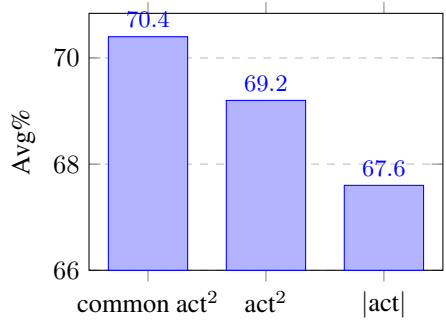
\begin{wrapfigure}[11]{r}{0.45\textwidth}
\begin{center}
\vspace{-16mm}
\begin{tikzpicture}
\begin{axis}[
    ybar,
    bar width=30pt,
    width=\linewidth,
    height=5cm,
    ymin=66,
    ylabel={Avg\%},
    symbolic x coords={common act$^2$, act$^2$, $|$act$|$},
    xtick=data,
    nodes near coords,
    nodes near coords align={vertical},
    every node near coord/.append style={font=\small},
    enlarge x limits=0.25,
    ymajorgrids=true,
    grid style=dashed,
]
\addplot coordinates {(common act$^2$, 70.4) (act$^2$, 69.2) ($|$act$|$, 67.6)};
\end{axis}
\end{tikzpicture}
\end{center}
\vspace{-5mm}
\caption{ \commonact~outperforms both \actsquare~and $|$act$|$ (Qwen 2.5-0.5B at a 35\% pruning ratio).}
\label{fig:interprunetech}
\end{wrapfigure}

\vspace{-2mm}
\section{Ablation Study}
\vspace{-2mm}
In our ablation study, we try to further four questions: i) $\mathcal{Q}_1$: how effective is \commonact? ii) $\mathcal{Q}_2$: how to trade off \vocabprune~and \interprune? iii) $\mathcal{Q}_3$: how much calibration data? 

\paragraph{Answer to $\mathcal{Q}_1$: effectiveness of \commonact.}

We compare our novel \commonact~method with \actsquare. To isolate the effect of the intermediate pruning method, we prune rare vocabulary for all models, then apply the specified intermediate pruning method. We also test another commonly used method, $|$act$|$, which is similar to \actsquare~but uses the summed \textit{absolute} activations instead of squared activations. Our results are summarized in Figure \ref{fig:interprunetech}. We find that \commonact~achieves the highest mean performance compared to our baselines.

\begin{wrapfigure}[14]{r}{0.5\textwidth}
\vspace{-10mm}
\begin{center}
\begin{tikzpicture}
\begin{axis}[
    width=\linewidth,
    height=0.35\textwidth,
    xlabel={\# Samples},
    ylabel={Avg\%},
    xmode=log,
    log basis x={2},
    xtick={8,16,32,64,128,256,512,1024},
    xticklabels={8,16,32,64,128,256,512,1024},
    ymin=66, ymax=72,
    grid=both,
    grid style={dashed,gray!30},
    mark size=2pt,
    line width=1pt
]
\addplot+[mark=o,blue] coordinates {
    (8,66.8)
    (16,70.9)
    (32,71.2)
    (64,70.1)
    (128,69.7)
    (256,70.4)
    (512,70.4)
    (1024,70.8)
};
\end{axis}
\end{tikzpicture}
\end{center}
\vspace{-6mm}
\caption{Downstream performance by calibration dataset size. Although all benchmarks used 256 calibration samples, 16 samples is sufficient.}
\label{fig:calibration}
\end{wrapfigure}
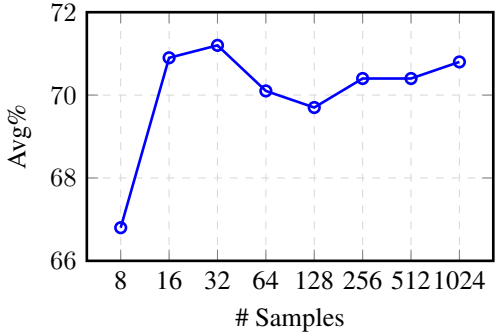

\paragraph{Answer to $\mathcal{Q}_2$: vocabulary-intermediate tradeoff}

\modelname~has two hyperparameters: the new vocabulary size $V'$ and the new intermediate size $I'$. These can be adjusted to produce many configurations at a given pruning ratio. We test different configurations of Qwen 2.5-0.5B at 35\% pruning. Our results are in Table \ref{tab:tradeoff}. Note that $V'=151936,I'=2048$ is identical to \actsquare. Despite \actsquare's simplicity, it achieves better performance than our baselines even without \vocabprune. However, the best result is achieved with a combination of both, validating \modelname's methodology.

\begin{table}%[htbp]
\vspace{-4mm}
\caption{Multiple values of $V'$ and $I'$ over a fixed pruning ratio for Qwen 2.5-0.5B.}
\begin{center}
\begin{adjustbox}{width=0.6\textwidth}
\begin{tabular}{c|cc|c|cc}
\hline
& \textbf{V'} & \textbf{I'} & \textbf{Ratio} & \textbf{Avg} & \textbf{Avg\%} \\
\hline

Dense & 151936 & 4864 & 0.00\% & 50.2 & 100.0 \\
\hline
\actsquare & 151936 & 2048 & 36.84\% & 31.6 & 63.0 \\\cdashline{1-6}
\multirow{8}{*}{\modelname} & 131584 & 2304 & 37.04\% & 31.8 & 63.3 \\
&111104 & 2560 & 37.45\% & 33.1 & 66.1 \\
&90752  & 2944 & 36.23\% & 34.2 & 68.2 \\
&70400  & 3200 & 36.44\% & 34.5 & 68.9 \\
&49536  & 3456 & 37.04\% & \textbf{35.3} & \textbf{70.4} \\
&29568  & 3712 & 37.25\% & 34.7 & 69.2 \\
&9088   & 3968 & 37.65\% & 32.4 & 64.7 \\\hline
\end{tabular}\label{tab:tradeoff}
\end{adjustbox}
\end{center}
\vspace{-4mm}
\end{table}

\paragraph{Answer to $\mathcal{Q}_3$: calibration data size.}

We perform ablations over the number of calibration samples, with our results in Figure \ref{fig:calibration}. \modelname~is highly robust to sample count, and similar performance can be achieved with just 16 samples, implying that the pruning time can be reduced further.

%\vspace{-2mm}
\section{Conclusion}
%\vspace{-3mm}
We propose \modelname, a training-free pruning method combining vocabulary and FFN pruning under the common-token distribution. Experiments show that it achieves robust performance across model scales, offering smooth degradation, strong efficiency, and broad deployment compatibility. In future works, we plan to address the discrepancy in throughput between our method and ShortGPT/LaCo, closing the gap between width and depth pruning.

%\vspace{-1mm}
%\paragraph{LLM Usage Disclosure.}
%We used GPT-5 to assist with language polishing of the manuscript. No parts of the methodology, %experimental design, or results were generated by an LLM.

\bibliography{iclr2026_conference}
\bibliographystyle{iclr2026_conference}

\appendix
\section{Appendix}

\subsection{Parameter Distribution Across Other Model Families}
\label{appendix:distribution}

Figures \ref{fig:distributionllama} and \ref{fig:distributiongemma} provide parameter distributions for the LLaMA 3 and Gemma 3 model families, respectively. We see that these models follow the same trend as Qwen 2.5 where SLMs have a higher proportion of vocabulary parameters, corroborating our theoretical analysis.

\begin{figure}
\begin{center}
\begin{minipage}{0.48\textwidth}
\vspace{-7mm}
\begin{tikzpicture}
\begin{axis}[
    ybar stacked,
    bar width=18pt,
    width=0.84\textwidth,
    height=0.7\textwidth,
    ymin=0, ymax=1,
    ylabel={Proportion},
    symbolic x coords={1B,3B,8B,70B,405B},
    xtick=data,
    x tick label style={rotate=45,anchor=east},
    legend style={at={(0.5,-0.25)},anchor=north,legend columns=-1},
    enlarge x limits=0.15,
    axis line style={draw=none},
    ymajorgrids=true,
    grid style=dashed,
]
\addplot+[fill=lightblue] coordinates {(1B,0.213) (3B,0.123) (8B,0.131) (70B,0.030) (405B,0.010)};
\addplot+[fill=lightred] coordinates {(1B,0.652) (3B,0.658) (8B,0.702) (70B,0.799) (405B,0.813)};
\addplot+[fill=lightgreen] coordinates {(1B,0.136) (3B,0.219) (8B,0.167) (70B,0.171) (405B,0.177)};
\legend{Vocab,FFN,Attention}
\end{axis}
\end{tikzpicture}
\vspace{-3mm}
\caption{Llama 3 Parameter Distribution}
\label{fig:distributionllama}
\end{minipage}%
\hfill
\begin{minipage}{0.48\textwidth}
\centering
\vspace{-7mm}
\begin{tikzpicture}
\begin{axis}[
    ybar stacked,
    bar width=18pt,
    width=0.84\textwidth,
    height=0.7\textwidth,
    ymin=0, ymax=1,
    ylabel={Proportion},
    symbolic x coords={270M,1B,4B,12B,27B},
    xtick=data,
    x tick label style={rotate=45,anchor=east},
    legend style={at={(0.5,-0.25)},anchor=north,legend columns=-1},
    enlarge x limits=0.15,
    axis line style={draw=none},
    ymajorgrids=true,
    grid style=dashed,
]
\addplot+[fill=lightblue] coordinates {(270M,0.626) (1B,0.302) (4B,0.173) (12B,0.086) (27B,0.052)};
\addplot+[fill=lightred] coordinates {(270M,0.264) (1B,0.621) (4B,0.689) (12B,0.722) (27B,0.796)};
\addplot+[fill=lightgreen] coordinates {(270M,0.110) (1B,0.077) (4B,0.138) (12B,0.193) (27B,0.152)};
\legend{Vocab,FFN,Attention}
\end{axis}
\end{tikzpicture}
\vspace{-3mm}
\caption{Gemma 3 Parameter Distribution}
\label{fig:distributiongemma}
\end{minipage}
\end{center}
\end{figure}

\subsection{Evaluation Methodology}
\label{appendix:eval}
Downstream evaluations were conducted using the LM-Evaluation-Harness \citep{eval-harness}, specifically  \texttt{lm-eval 0.4.8}. The evaluation details for each benchmark are in Table \ref{tab:eval}.

\begin{table}%[htbp]
\caption{Evaluation methodology for each benchmark.}
\begin{center}
\begin{adjustbox}{width=0.7\textwidth}
\begin{tabular}{c|ccc}
\hline
\textbf{Benchmark} & \textbf{n-shot} & \textbf{Type} & \textbf{Metric} \\
\hline
\textbf{MMLU}        & 0 & multiple-choice & acc \\
\textbf{HellaSwag}   & 0 & multiple-choice & acc\_norm \\
\textbf{WinoGrande}  & 0 & multiple-choice & acc \\
\textbf{ARC-C}       & 0 & multiple-choice & acc\_norm \\
\textbf{ARC-E}       & 0 & multiple-choice & acc\_norm \\
\textbf{PIQA}        & 0 & multiple-choice & acc\_norm \\
\textbf{GSM8K}       & 5 & generative & strict\_match \\
\hline

\end{tabular}\label{tab:eval}
\end{adjustbox}
\end{center}
\end{table}

\subsection{Pruning Hyperparameters}
\label{appendix:hyperparameters}
We provide $V'$ and $I'$ for all our main results in Table \ref{tab:hyperparameters}. These hyperparameters were found by sweeping over all possible configurations for the given pruning ratio, similarly to Table \ref{tab:tradeoff}.

\begin{table}%[htbp]
\caption{\modelname~pruning hyperparameters for all main results.}
\begin{center}
\begin{adjustbox}{width=0.5\textwidth}
\begin{tabular}{c|c|cc}
\hline
& \textbf{Ratio (\%)} & \textbf{V'} & \textbf{I'} \\
\hline

\multirow{4}{*}{\textbf{Qwen 2.5--0.5B}}
& 0.00  & 151936 & 4864 \\
& 10.00 & 99968 & 4736 \\
& 20.00 & 49536 & 4736 \\
& 35.00 & 49536 & 3456 \\
\hline

\multirow{4}{*}{\textbf{LLaMA 3.2--1B}}
& 0.00  & 128256 & 8192 \\
& 10.00 & 67968 & 8192 \\
& 20.00 & 56704 & 7168 \\
& 35.00 & 33792 & 5760 \\
\hline

\multirow{4}{*}{\textbf{Gemma 3--1B}}
& 0.00  & 262144 & 6912 \\
& 10.00 & 174592 & 6912 \\
& 20.00 & 86912 & 6912 \\
& 35.00 & 95232 & 5120 \\
\hline

\multirow{4}{*}{\textbf{LLaMA 3.1--8B}}
& 0.00  & 128256 & 14336 \\
& 10.00 & 73216 & 13440 \\
& 20.00 & 67328 & 11520 \\
& 35.00 & 67840 & 8448 \\
\hline

\multirow{4}{*}{\textbf{LLaMA 3.1--70B}}
& 0.00  & 128256 & 28672 \\
& 10.00 & 112384 & 25216 \\
& 20.00 & 111872 & 21632 \\
& 35.00 & 110976 & 16256 \\
\hline

\end{tabular}\label{tab:hyperparameters}
\end{adjustbox}
\end{center}
\end{table}

\subsection{Inference Settings}
\label{appendix:inference}
Settings: i) \textbf{Text classification:} When running our pruned models on our downstream performance benchmarks, we record the maximum memory usage as well as the throughput, measured in number of questions per second. For this test, we use the HellaSwag benchmark, as it is the longest test in our benchmark suite, which allows us to better discriminate between the methods. We test on a larger model (LLaMA 3.1-8B) using a 3-run average, again to better discriminate between methods. All models are loaded in 16-bit precision on a single A100-80GB GPU, with a batch size of 256. ii) \textbf{Text generation:} We use the vLLM library \citep{kwon2023efficient} to test the memory reduction and inference speedup of our method. As mentioned before, because SliceGPT and 2SSP are incompatible with vLLM, we omit it from our tests. Similarly to the text classification test, we use a 3-run average of LLaMA 3.1-8B in 16-bit precision on a single A100-80GB GPU, but with a batch size of 1 instead, as this is a more realistic workload for on-device text generation. Tests are conducted with 128 input tokens and 128 output tokens.

\end{document}